\documentclass[final,5p,times,twocolumn]{elsarticle}

\usepackage[utf8]{inputenc}
\usepackage{newunicodechar}
\newunicodechar{ℓ}{l}

\usepackage{lineno}
\usepackage{graphicx}
\usepackage{amssymb}
\usepackage{pdflscape}
\usepackage[english]{babel}
\usepackage[dvipsnames,svgnames,x11names]{xcolor}
\usepackage{booktabs,tabularx}
\usepackage{multirow}
\usepackage{subfig} 
\usepackage{hyperref}
\usepackage{amsmath}
\usepackage{commath}
\usepackage[english]{babel}
\usepackage[ruled]{algorithm2e}
\SetEndCharOfAlgoLine{}
\usepackage[leftmargin=6em,rightmargin=12em,indentfirst=false]{quoting}
\usepackage{siunitx}

\usepackage[final]{changes}
\usepackage{float}

\usepackage{lineno}
\usepackage{tikz}
\usepackage[export]{adjustbox}

\usepackage{graphicx}
\usepackage[utf8]{inputenc}
\usepackage[export]{adjustbox}
\usepackage{wrapfig}
\usepackage{dcolumn}

\makeatletter
\newcolumntype{T}[3]{>{\textfont0=\the@{#1}{#2}{#3}}c<{\DC@end}}
\makeatother

\usepackage{pgfplots}
\pgfplotsset{width=10cm,compat=1.9}

\usepackage{array}

\newcolumntype{L}[1]{>{\raggedright\let\newline\\\arraybackslash\hspace{0pt}}m{#1}}
\newcolumntype{C}[1]{>{\centering\let\newline\\\arraybackslash\hspace{0pt}}m{#1}}
\newcolumntype{R}[1]{>{\raggedleft\let\newline\\\arraybackslash\hspace{0pt}}m{#1}}

\usepackage{subfiles}

\usepackage{todonotes}

\journal{ASHRAE Transactions}
\begin{document}
	
\begin{frontmatter}

\title{Gradient boosting machines and careful pre-processing work best: ASHRAE Great Energy Predictor III lessons learned}

\author{Clayton Miller$^{*}$, Liu Hao, Chun Fu}

\address{Department of the Built Environment, National University of Singapore (NUS), Singapore}

\address{$^*$Corresponding Author: clayton@nus.edu.sg, +65 81602452}

\begin{abstract}
The ASHRAE Great Energy Predictor III (GEPIII) competition was held in late 2019 as one of the largest machine learning competitions ever held focused on building performance. It was hosted on the Kaggle platform and resulted in 39,402 prediction submissions, with the top five teams splitting \$25,000 in prize money. This paper outlines lessons learned from participants, mainly from teams who scored in the top 5\% of the competition. Various insights were gained from their experience through an online survey, analysis of publicly shared submissions and notebooks, and the documentation of the winning teams. The top-performing solutions mostly used ensembles of Gradient Boosting Machine (GBM) tree-based models, with the LightGBM package being the most popular. The survey participants indicated that the preprocessing and feature extraction phases were the most important aspects of creating the best modeling approach. All the survey respondents used Python as their primary modeling tool, and it was common to use Jupyter-style Notebooks as development environments. These conclusions are essential to help steer the research and practical implementation of building energy meter prediction in the future.
\end{abstract}



\end{frontmatter}


\section{Introduction}

Machine learning (ML) for building energy prediction is a rich research community with hundreds of influential publications~\cite{Amasyali2018-wj}. However, a fundamental challenge in the literature is a lack of comparability of prediction techniques~\cite{Miller2019-sg}, despite previous efforts at benchmarking~\cite{Granderson2015-ms,Granderson2016-wq}. Machine learning competitions provide a comparison of techniques through the crowdsourcing and benchmarking of various combinations of models with a reward for the most objectively accurate solution.

ASHRAE has hosted three machine learning energy prediction competitions since 1993. The first two competitions were named the Great Energy Predictor Shootouts I and II held in 1993 and 1994. In the first competition, each contestant was given a four-month data set to predict building energy use and insolation data for the next two months, and the final model, which was developed using Bayesian nonlinear modeling and artificial neural networks, was found to be the most effective and accurate~\cite{Kreider1994-dn}. While in the second competition, entrants were asked to predict the intentionally removed portion of data based on the existing building's pre-retrofit and post-retrofit data~\cite{Katipamula1996-et}. Derived from the submissions that met the requirements, neural networks had shown to be the most accurate model overall, while cleverly assembled statistical models were additionally found to be even more accurate than neural networks in some fields~\cite{Haberl1998-du,Haberl1999-tq}. Despite the passage of time, the contents of these two competitions are still being investigated and used for references. 

After more than two decades, the third competition, titled the \emph{ASHRAE Great Energy Predictor III (GEPIII)}, was initiated at the ASHRAE Winter Conference in Chicago in January 2018. After getting the approval and financial sponsorship from the ASHRAE Research Activities Committee (RAC), the competition was officially launched\footnote{\url{https://www.kaggle.com/c/ashrae-energy-prediction}} on 5 October 2019, and it ended on 19 December 2019~\cite{Miller2020-qw}. 
The context of the GEPIII competition was whether energy-saving building retrofitting could help to improve energy efficiency~\cite{Grillone2020-fm}. The datasets utilized in the contest were collected from around 1,440 buildings from 16 sites worldwide, of which 73\% were educational, and the other 27\% were municipal and healthcare facilities. The energy meter readings of these buildings from January 2016 to December 2018 were combined to form the dataset. Based on this context, the participants were challenged to create a counterfactual model to estimate the building's pre-renovation energy usage rate in the post-renovation period~\cite{Miller2020-qw}. The final ranking of contestants was determined by the Private Leaderboard (PLB) scores, and the top five performers were awarded monetary prizes. 

One key output of the competition was learning from the contestants’ solutions and understanding the general nature of what makes the best machine learning solution for long-term building energy prediction. This paper outlines a post-competition survey used to capture the demographics, best practices, and lessons learned from a subset of the top-performing teams.

\section{Methodology}
\label{sec:methodology}

The primary goal of this analysis was to investigate the demographics and machine learning strategy preferences of participants from the top teams in the GEPIII competition. The primary component of this methodology was a survey composed of a segment that included questions regarding respondents' background information and another with queries regarding the final solutions they submitted. The additional data collection process was done by analyzing the publicly available analysis notebooks posted as part of the competition and the submissions and interviews of the top five winning teams.

The first portion of the web-based survey gathered some basic information about the contestants, such as age, gender, educational information, current job fields, and work experience. The second portion of the survey was designed to gather information regarding their participation experience. It can be further broken down into three subsections. The first subsection is primarily intended to collect information about how contestants arrived at their final solutions, such as what programming languages they utilized, what platforms they selected most to run their codes and the methods and algorithms employed in each phase of building machine learning models. Furthermore, the participants were asked to express their opinions on the significance of the five steps machine learning workflow in the second subsection. In the last subsection, contestants are asked to provide their feedback on the competition by commenting on what parts they liked or disliked. These insights were designed to help the organizers understand what competition mechanisms the contestants prefer, allowing improvements for future events. 

This paper aims to outline the collection of insights targeting the teams that scored in the top 5\% (180 teams who earned gold or silver medals) of the competition. This process seeks to characterize the insights and best practices of the contestants and teams that created solutions with the highest performance. Towards this effort, the survey was sent to teams from the top-performing competition participants from May to August 2021. We received responses from 27 individuals that included the collective insights from 50 contestants, with team members of respondents contained for non-demographic questions. We included the data from publicly available online solutions posted by another 34 teams, including the top five winning teams for the tools and modeling analysis. Most of the data were collected from teams in the top 5\% (including 90\% of the survey data).

\begin{figure*}[!h]
    \centering
    \includegraphics[width=0.95\linewidth]{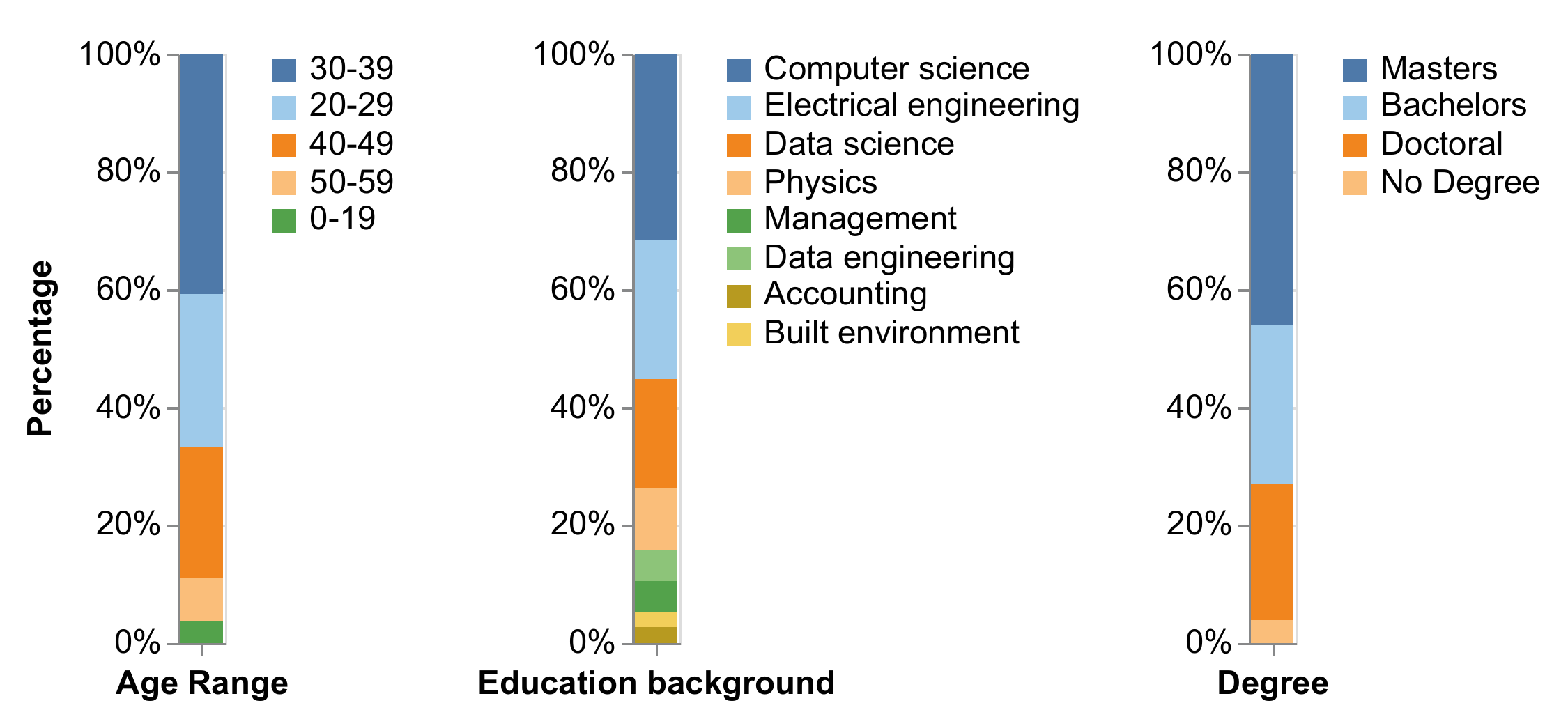}
    \caption{Overview of selected demographic information about the contestants, including their age range (left), educational background (middle), and type of highest degree awarded (right).}
    \label{fig:fig1}
\end{figure*}

\section{Results}
\label{sec:results}

\subsection{Demographics of contestants}

One key focus of the competition was to work towards the better exchange of ideas between the building and construction industry and the fast-growing data science community. The first analysis from the survey data was to understand what backgrounds the contestants generally had. Figure \ref{fig:fig1} illustrates the high-level demographics of the survey respondents.

The age range results showed that 40\% of the respondents are between the ages of 30 and 39 and a further 23\% are between 20-29. These results compare well to a larger-scale generic survey collected by Kaggle in 2021, the composition of this sample group was quite close to that result, namely that around 80\% of Kaggle data scientists aged between 22 and 44 years old (https://www.kaggle.com/c/kaggle-survey-2021). The educational background of the contestants is dominated by \emph{Computer Science}, \emph{Electrical Engineering}, and \emph{Data Science}, which is also similar to the make-up of the general ecosystem and to be expected in the context of machine learning expertise. A vast majority of the contestants have higher education degrees, with over 75\% being either Master’s or Doctoral Post-graduate education.

\subsection{Most popular and effective tools and modelling approaches}
\begin{figure*}[!h]
    \centering
    \includegraphics[width=0.95\linewidth]{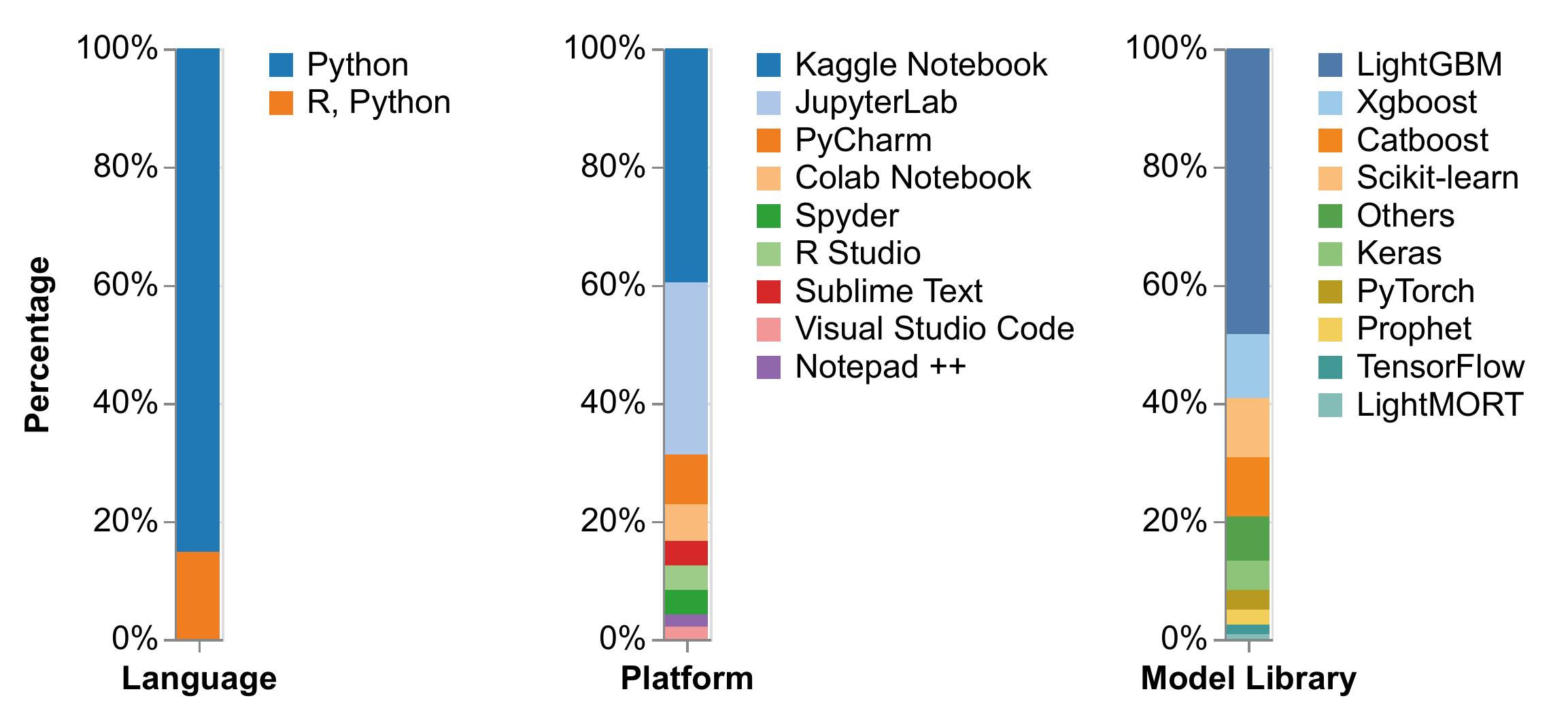}
    \caption{Overview of the tools and modeling approaches the contestant used, including programming language (left), development platform/environment (middle), and libraries of models used for their solution (right).}
    \label{fig:fig2}
\end{figure*}

The next focus of the survey was to characterize the contestants' tools and models to understand what works well for the building energy prediction context. Figure \ref{fig:fig2} outlines the survey results for the programming language, model development tool/environment/platform, and the model library used.

The most conclusive result from the survey was that the open-source Python programming language came out on top in terms of tools used, with all survey respondents indicating its use in their solutions. About 18\% of the contestants also used the R programming language in parallel with their Python-based solution. In terms of the platforms used to develop the solutions, most contestants used some form of a \emph{notebook} format. Notebooks in the context of data and computer science are web-based interfaces with a series of cells containing snippets of code or explanatory text interspersed and are meant to be executed one cell at a time for a user to replicate a process of analysis one step at a time. Kaggle, Jupyter, and Colab notebooks are similar in their functionality and were among the most popular solutions used.

One of the most interesting results of the competition and survey was the dominance of \emph{Gradient Boosting Machines} (GBM), also known as Gradient Boosting Decision Trees (GBDT), as the modeling technique most used in high-performance solutions. This result correlates well with the overview analysis of the top five winning solutions and their associated documentation~\cite{Miller2020-qw} and previous work~\cite{Touzani2018-sa}. Several Python packages are specialized in this modeling category, such as \emph{LightGBM}~\cite{Ke2017-ya}, \emph{XGBoost}~\cite{Chen2016-vx}, and \emph{CatBoost}~\cite{Prokhorenkova2018-qk} that show up as the top three libraries used by the survey respondents and the winning solutions. The popularity of these libraries can be compared to the use of deep learning libraries such as \emph{TensorFlow}~\cite{Abadi2016-ic} or \emph{Keras}~\cite{Gulli2017-wt}. All packages used are open-source and predominantly implemented using the Python programming language.

\subsection{Relative importance of machine learning process steps}
The final portion of the survey focuses on the steps that the competition participants took to develop their solutions. The machine learning process was divided into five phases, and the participants were asked about their perception of the importance of each phase towards the success of their solution. The phases included the pre-processing and feature engineering steps in the early stages of the machine learning process in which the training data are prepared for the models. These phases include the removal of anomalous outlier behavior that could be unhelpful in training the model and converting the raw time-series data into various input features that can be used to train the model. The modeling phase includes selecting, training, and tuning the machine learning models themselves. The post-processing and validation phases occur after modeling to combine and weight the outputs of ensembles of various models and compare the performance of their solution to the training and public data sets that the contestants found online (public leaderboard leakage data). 

\begin{figure*}[!h]
    \centering
    \includegraphics[width=0.95\linewidth]{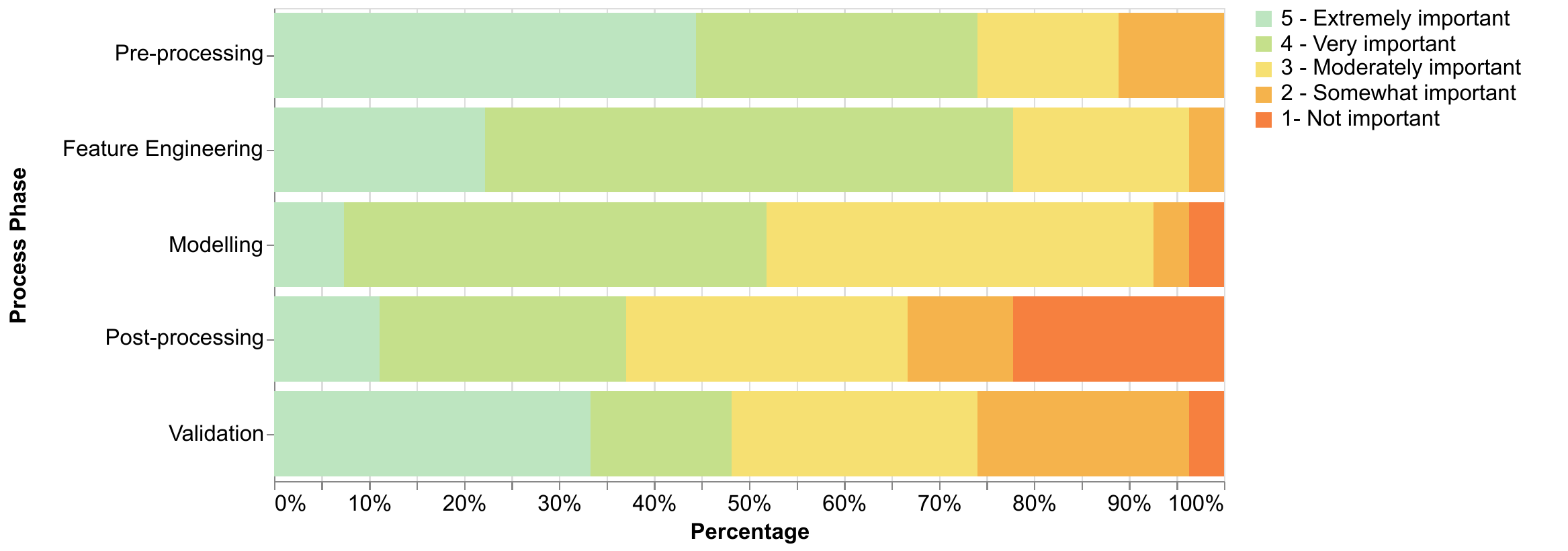}
    \caption{Overview of the importance of each machine learning phase towards the success of solutions in the top 5\% of the GEPIII competition.}
    \label{fig:fig3}
\end{figure*}

Figure \ref{fig:fig3} shows an overview of the importance of the phases according to the survey respondents. The \emph{Pre-processing} and \emph{Feature engineering} phases have the highest total percentage of respondents, considering them to be either \emph{Extremely} or \emph{Very important}, with both being over 70\% total for those two categories. This result supports the insight gained from the top five winning solution winners when they stated that some of the most important work they did was to carefully remove outliers and anomalous behavior from the training data. This effort increased the generalizability of their modeling solutions on the two types of test data sets. The modeling and post-processing steps were found to be relatively less important than the others, thus showing that time spent optimizing hyperparameters and the structure of ensembles of models was considered less crucial to the success of the solution. 

\section{Discussion}
\label{sec:discussion}

\subsection{Comparison to the Great Energy Predictor I and II competitions}
At the launch of GEPIII, about 25 years had passed since the Great Energy Predictor I and II competition of the early 1990s. During the period of these older competitions, there was an emergence of data-driven modeling, and several modeling techniques that are still popular today, such as neural networks and Bayesian models, emerged. The winning solutions of those older competitions included various configurations of these types of models~\cite{Chonan1996-rz,Feuston1994-mm,Iijima1994-gj,Ohlsson1994-vl,Stevenson1994-ps}. With the GEPIII competition, there was a concerted use of newer types of machine learning models such as Gradient Boosting Machines (GBM) that have been invented in the period since the early 1990s. Techniques such as ensembling of large networks of models also emerged as a method that showed progress due to the much larger number of meters included in the modern version of the competition as well as the increase in data processing capabilities.  Another significant difference was that the generation of content on a modern machine learning competition platform enables a large number of solutions beyond the top five to be shared openly. This crowdsourced content included numerous examples of the machine learning process flow that are useful for beginners. 

\subsection{Gradient Boosting Machine (GBM) tree-based models were popular}

The results of the survey combined with presentations and discussions by the winning teams highlighted that \emph{Gradient Boosting Machines (GBM)}, and tree-based models in general, were practically useful and popular for the tabular time-series data context of the competition. These results align with studies from other domains that show these models often being the most effective for time-series forecasting~\cite{Elsayed2021-zc,Manibardo2021-vj}. GBM’s are more suitable for tabular datasets (also known as structured data) than alternatives like deep learning (DL), especially in the tasks of classification and regression~\cite{Shwartz-Ziv2022-sn}. In ML competitions in general, the winning solutions for competitions with tabular data are usually GBM models, and those for unstructured data (such as image and audio data) are mostly DL methods. In addition, DL usually requires a larger dataset due to more parameters to be trained, while the training data of GEPIII is a medium-sized dataset (~700 MB). These characteristics might explain why solutions in the GEPIII are mostly GBM-based. This result might be good for practitioners as decision-tree model types are easier to implement, more interpretable, and require less data to be effective than other approaches. GEPIII gives a different perspective to previous work that compares tree-based models for this context~\cite{Ahmad2017-jz}. 

\subsection{Pre-processing and feature engineering phases were important}

The emergence of the pre-processing and feature extraction phases of the machine learning process for this context as the most important towards the solution’s success was an interesting highlight of the survey. This result corresponded well also with the interviews and analysis of the top five winning team solutions. These steps often require domain expertise in understanding what type of behavior should or should not be included in the training models. The fact that these steps emerged as important reinforces that a solid understanding of building energy is essential for the best implementation of data-driven modeling methods in practice. In addition, there are innovations in automated discord and change-point detection that could further enhance the accuracy of prediction models are reduce the manual implementation cost~\cite{Park2020-zx,Touzani2019-sz}. 

\subsection{What are the opportunities for improvement of machine learning for energy prediction?}

The GEPIII competition established a foundation for understanding the limitations of machine learning for long-term, whole-building energy meter prediction. Due to the large number of data science experts all competing against each other for the most accurate solutions, the winners’ submissions can be considered near-optimal for the context and data provided. However, there are numerous paths forward for innovation through the inclusion of new training data such as occupancy-based energy use proxies such as WiFi~\cite{Nweye2020-yz,Zhan2021-jy} and online digital sources~\cite{Lu2021-xm,Fu2021-fw} as well as methods to process them such as pattern classification~\cite{Dong2021-xb} and probabilistic approaches~\cite{Roth2021-be}. There is emerging work in meshing various IoT and spatial data sources from buildings towards potential in multiple applications~\cite{Miller2021-bm}, including energy prediction. Recent work to characterize the GEPIII competition error also shows which building and meter types need innovation to improve accuracy~\cite{Miller2021-bf}.

\subsection{How do we get more built environment professionals into data science?}

A prominent result of the demographics portion of the survey was that there were few built environment-related professionals in the competition. The development of this technology in practice will likely not come from more accurate models but from adopting skills in the industry to implement and improve such solutions. The top winning solutions were open-sourced for the community to use and learn from. However, there were also a large number of machine learning workflows that were shared that can help \emph{beginners} in understanding how to use modern data science techniques to process the competition data set. These notebooks combined with an even larger open dataset~\cite{Miller2020-yc} that was released after the competition are possible catalysts for more built environment experts to learn the skills that help in the innovation of machine learning for this context. 

\subsection{Limitations}

The analysis outlined in this study has several limitations related to the size of the survey data collection and the nature of machine learning competitions compared to more conventional research. The number and diversity of survey respondents were lower than optimal due to the amount of time that had passed since the competition was held and when the survey was released. This situation introduces uncertainty and bias based on the survey respondents who participated in the data collection process. Another aspect of this situation is that many of the non-machine learning experts in the competition likely scored further down the leaderboard rankings. The survey doesn’t represent those participants that are more likely to have building industry experience and are just starting to learn data science skills. 

Another limitation is the comparison of these results with the wider research community literature. Machine learning competitions are excellent at getting a large number of people to compare approaches and techniques in a scaled way, but they are limited in how complex the objective can be. In the GEPIII competition, long-term hourly energy meter prediction was the objective, therefore the results of the most accurate solutions could be highly focused on this particular objective. However, there are other types of machine learning objectives that may have complexities that require more advanced techniques to create the optimal solutions~\cite{Gao2021-sn}. Future work should focus on answering further questions about which types of techniques are most effective in a more extensive range of contexts.

\section{Conclusion}
\label{sec:conclusion}

This paper gives an overview of the results of a survey focused on characterizing the lessons learned from the ASHRAE GEPIII machine learning competition hosted in 2019. This analysis found several interesting insights into the best practices of using machine learning to predict the long-term whole building energy consumption across a large number of buildings. The results illustrated the dominance of open-source tools such as \emph{Python} and libraries such as \emph{LightGBM} in the development of the most cutting-edge machine learning solutions. The results highlighted critical areas of focus in extending and building upon the competition, such as improving the breadth of behavior captured in training data by including additional sources. The analysis also provided context on how to better extend data science skills to professionals in the built environment through the use of open datasets and libraries that can be used for beginners to learn how to use the tools and for non-domain data science experts to understand the nature of built environment data.



\section*{Acknowledgements}
This analysis is possible due to the GEPIII competition planning and operations committees. The technical committee to be acknowledged includes (alphabetical order) Anjukan Kathirgamanathan, June Young Park, Pandarasamy Arjunan, and Zoltan Nagy. Planning committee members to be acknowledged are Anthony Fontanini, Chris Balbach, Jeff Haberl, and Krishnan Gowri. The ASHRAE organization is recognized for providing support for the competition prize money and the Kaggle platform for hosting GEPIII as a non-profit competition. In addition, the authors would like to thank those who assisted in collecting and releasing the BDG2 data set, including Bianca Picchetti, Brodie Hobson, Forrest Meggers, Paul Raftery, and Zixiao Shi. The Singapore Ministry of Education (MOE) provided support for the development and implementation of this research through the \emph{Temporal Mining of Energy and Indoor Environmental Quality Data from Buildings} (R296000181133) Project.

\bibliographystyle{model1-num-names}
\bibliography{references}

\end{document}